\documentclass[letterpaper, 10 pt, conference]{ieeeconf}  

\IEEEoverridecommandlockouts                              

\usepackage{graphicx} 
\usepackage{times} 
\usepackage{amsmath} 
\usepackage{amssymb}  
\usepackage{bbold}
\usepackage{url}
\usepackage{cite}
\usepackage{enumerate}
\usepackage{booktabs}
\usepackage{tabularx}
\usepackage{multirow}
\usepackage{pifont}
\usepackage[table,xcdraw]{xcolor}
\usepackage{xspace}
\usepackage{float} 

\makeatletter
\let\NAT@parse\undefined
\makeatother
\usepackage[pagebackref=false,breaklinks=true,letterpaper=true,colorlinks,bookmarks=false]{hyperref}

\definecolor{baselinecolor}{gray}{.9}

\newcommand{\fadedtext}[1]{\textcolor{gray}{#1}}

\newcolumntype{x}[1]{>{\centering\arraybackslash}p{#1pt}}
\newcolumntype{y}[1]{>{\raggedright\arraybackslash}p{#1pt}}
\newcolumntype{z}[1]{>{\raggedleft\arraybackslash}p{#1pt}}
\newlength\savewidth

\renewcommand{\paragraph}[1]{\vspace{1.25mm}\noindent\textbf{#1}}

\newcommand{\cmark}{\ding{51}}%
\newcommand{\xmark}{\ding{55}}%

\makeatletter
\DeclareRobustCommand\onedot{\futurelet\@let@token\@onedot}
\def\@onedot{\ifx\@let@token.\else.\null\fi\xspace}
\def\eg{\emph{e.g}\onedot} 
\def\ie{\emph{i.e}\onedot}

\title{\LARGE \bf
Rethinking Imitation-based Planner for Autonomous Driving
}

\author{Jie Cheng$^{1}$, Yingbing Chen$^{1}$, Xiaodong Mei$^{1}$, Bowen Yang$^{1}$, Bo Li${^2}$ and Ming Liu$^{1,3}$
\thanks{$^{1}$Jie Cheng, Yingbing Chen, Xiaodong Mei and Bowen Yang are with the Hong Kong University of Science and Technology, Hong Kong SAR, China.\texttt{\{jchengai,ychengz,xmeiab,byangar\}@connect.ust.hk}}%
\thanks{$^{2}$Bo Li is with Lotus Technology Ltd. \texttt{libo@lotuscar.com.cn}}%
\thanks{$^{3}$Ming Liu is also with The Hong Kong University of Science and Technology (Guangzhou), Nansha, Guangzhou, China. \texttt{eelium@ust.hk}} %
}

\begin{document}

\maketitle
\thispagestyle{empty}
\pagestyle{empty}

\begin{abstract}

In recent years, imitation-based driving planners have reported considerable success. However, due to the absence of a standardized benchmark, the effectiveness of various designs remains unclear. The newly released nuPlan addresses this issue by offering a large-scale real-world dataset and a standardized closed-loop benchmark for equitable comparisons. Utilizing this platform, we conduct a comprehensive study on two fundamental yet underexplored aspects of imitation-based planners: the essential features for ego planning and the effective data augmentation techniques to reduce compounding errors. Furthermore, we highlight an imitation gap that has been overlooked by current learning systems. Finally, integrating our findings, we propose a strong baseline model—PlanTF. Our results demonstrate that a well-designed, purely imitation-based planner can achieve highly competitive performance compared to state-of-the-art methods involving hand-crafted rules and exhibit superior generalization capabilities in long-tail cases. Our models and benchmarks are publicly available. Project website \url{https://jchengai.github.io/planTF}.

\end{abstract}

\section{Introduction}

Learning-based planners are considered a potentially scalable solution for autonomous driving, supplanting traditional rule-based planners~\cite{chen2023e2e, hagedorn2023rethinking,cheng2022gpir}. 
This has sparked significant research interest in recent years. 
In particular, imitation-based planners~\cite{bansal2018chauffeurnet, zhou2021exploring, vitelli2022safetynet, scheel2022urban, cheng2022mpnp, pini2023safepathnet, huang2023gameformer, huang2023dipp, guo2023ccil} are reported to achieve notable success in simulations and real-world scenarios.
Nevertheless,  these planners are predominantly trained and evaluated in diverse custom conditions (\eg varying datasets, metrics, and simulation setups) owing to the absence of a standardized benchmark. Consequently, it becomes challenging to compare and summarize effective design choices for constructing practical learning-based systems. 

Recently, the release of the large-scale nuPlan~\cite{caesar2021nuplan} dataset, alongside a standardized simulation benchmark, has provided a new opportunity for advancing learned motion planners.
Enabled by this fresh benchmark, we conduct in-depth investigations on several common and critical 
yet not fully studied design choices of the learning-based planner, aiming to provide constructive suggestions for future research. 
This paper concentrates on two overarching and fundamental facets of the imitation-based planner: the requisite ego features for planning and the efficacious techniques of data augmentation. 

The majority of imitation-based planning models~\cite{zhou2021exploring, scheel2022urban,cheng2022mpnp,pini2023safepathnet,vitelli2022safetynet,huang2023gameformer,huang2023dipp} follow the success of prediction models and inherently incorporate the past trajectory of the autonomous vehicle (AV) as an input feature, though imitation learning (IL) has frequently been noted for its tendency to acquire shortcuts from historical observations~\cite{bansal2018chauffeurnet, muller2005offroad, wang2019monocular,wen2020fighting}. Our research reaffirms that the past motion of the AV leads to significant closed-loop performance degradation. 
The planner achieves enhanced performance by solely utilizing the AV's present state. 
Surprisingly, it attains better closed-loop performance purely using the AV's current pose (position and heading). 
This implies that additional kinematic attributes typically deemed crucial for planning, such as velocity, acceleration, and steering, lead to a performance decline. To gain deeper insights into this phenomenon, we perform a sensitivity analysis to assess the impact of the AV's states on the resulting trajectory. Our experiments reveal that the planner can learn to exploit shortcuts from its kinematic states, even when past motion data is absent. To mitigate this challenge, we introduced a straightforward yet highly effective attention-based state dropout encoder, enabling the planner that utilize kinematic states to achieve optimal overall performance.

Imitation learning is also known to have compounding errors~\cite{ross2011reduction}. Perturbation-based augmentations~\cite{bansal2018chauffeurnet,zhou2021exploring,vitelli2022safetynet} are a commonly employed strategy to instruct the planner on recovering from deviations. We conduct comprehensive experiments exploring various augmentation techniques, including history perturbation, state perturbation, and future correction. Additionally, we demonstrate the indispensability of proper normalization for the effectiveness of augmentation. Furthermore, we identify an ignored imitation gap within current learning frameworks and illustrate its potential impact.  

Finally, by combining our findings, we provide a pure learning-based baseline model that demonstrates strong performance against state-of-the-art competitors on our standardized nuPlan benchmark.  Our contributions are summarized as follows:
\begin{enumerate}
    \item We perform an in-depth investigation on necessary features for ego planning, yielding counter-intuitive results contrary to mainstream practices. Furthermore, we introduced an effective attention-based state dropout encoder that attains the highest overall performance. 
    \item We conducted a comprehensive array of experiments involving various augmentation techniques, thereby elucidating an effective strategy to mitigate compounding errors. Additionally, we identified an overlooked imitation gap in current learning frameworks.
    \item By combining our findings, we provide an open baseline model with strong performance. All our code, benchmarks, and models will be publicly released, as a reference for future research. 
\end{enumerate}

\section{Related Work}

\paragraph{Imitation-based planners} 
are highly favored among learning-based planners due to their ease of convergence and typical scalability with data. They can be categorized into two distinct groups based on their input types:	

1) \textit{End-to-end.} 
End-to-end (E2E) methods~\cite{zeng2019nmp,hu2022st-p3,hu2023uniAD,jiang2023vad,codevilla2019cilrs,chen2020lbc,chen2022lav,chitta2021neat,chitta2022transfuser,zhang2022mmfn, shao2023interfuser, jia2023think} directly produce future trajectories using raw sensor inputs. 
Leveraging the closed-loop CARLA benchmark~\cite{dosovitskiy2017carla} and the collaborative efforts of the open-source community, E2E methods have achieved remarkable advancements within a short span of time: evolving from initial basic CNN-based approaches (LBC~\cite{chen2020lbc}, CILRS~\cite{codevilla2019cilrs}) to encompass multi-modal fusion (Transfuser~\cite{chitta2022transfuser}, NEAT~\cite{chitta2021neat}, MMFN~\cite{zhang2022mmfn}, Interfuser~\cite{chitta2022transfuser}, ThinkTwice~\cite{jia2023think}), as well as incorporating integrated perception and planning strategies (LAV~\cite{chen2022lav}, ST-P3~\cite{hu2022st-p3}, VAD~\cite{jiang2023vad}). 
However, due to limitations posed by the simulated environment, these methods typically function at low vehicle speeds, and the behavior of the simulated traffic agents lacks realism and diversity. 
Emerging and intriguing research, such as data-driven traffic simulation~\cite{igl2022symphony,li2023scenarionet} and realistic sensor emulation~\cite{amini2022vista,wu2023mars}, holds the potential to mitigate these issues.	

2) \textit{Mid-to-mid.} 
These approaches~\cite{cheng2022mpnp,scheel2022urban,huang2023gameformer,hu2023hotplan,vitelli2022safetynet,pini2023safepathnet,bansal2018chauffeurnet,huang2023dipp,guo2023ccil,renz2023plant} utilize post-perception outcomes as input and can directly learn from recorded real-world data. 
Chauffernet~\cite{bansal2018chauffeurnet} introduces the synthesis of perturbed trajectories to mitigate covariate shift, a practice that becomes common in subsequent studies. 
\cite{zhou2021exploring} further augment the training data with on-policy rollouts. 
Several works have demonstrated the capability to operate real vehicles (SafetyNet~\cite{vitelli2022safetynet}, UrbanDriver~\cite{scheel2022urban}, SafetyPathNet~\cite{pini2023safepathnet}).
Many include a post-optimizer (DIPP~\cite{huang2023dipp}, GameFormer~\cite{huang2023gameformer}, hotplan~\cite{hu2023hotplan}, pegasus~\cite{xiimitation}) to enhance the planner's robustness. 
All the abovementioned methods except hotplan use AV's history motion. 
Our study focuses on this category and provides an in-depth investigation of several critical design choices based on standardized data and benchmarks. 

\paragraph{Beyond imitation.} 
Another line of research aims to overcome the inherent limitations~\cite{muller2005offroad, wang2019monocular, wen2020fighting} of pure imitation learning (IL), such as utilizing environmental losses~\cite{bansal2018chauffeurnet, zhou2021exploring}, integrating IL with reinforcement learning~\cite{lu2022imitation_not_enough, liu2022improved, huang2022efficient}, and incorporating adversarial training, also known as closed-loop training~\cite{baram2017end, bronstein2022hierarchical, couto2023hierarchical}. 
Our work shows that the pure IL-based planner has not reached its limit and can be significantly improved with appropriate design.  

\section{Rethink Imitaion-based Planner}

\begin{figure}
    \centering
    \includegraphics[width=0.65\linewidth]{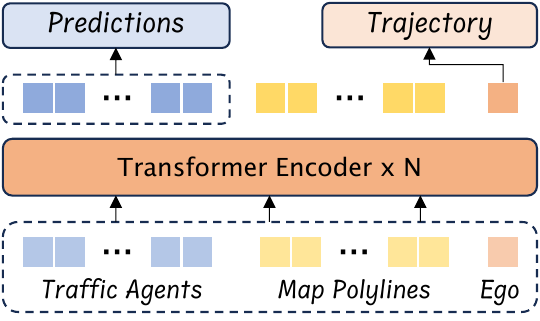}
    \vspace{-0.5em}
    \caption{A brief overview of our baseline model. Agents, map, and ego features are separately encoded and then concatenated, which are subsequently processed by a stack of transformer encoder layers. The baseline model jointly predicts traffic agents and plans for ego vehicle at the scene level.}
    \label{fig:baseline_model}
\vspace{-0.5em}
\end{figure} 

We consider the task of urban navigation employing a learned planner, trained by imitating the expert trajectory from the dataset. 
At each planning iteration, the planner receives various inputs, such as tracking data of surrounding objects up to a 2-second historical window, the current and past kinematic states of the ego vehicle, information about traffic lights, high-definition (HD) maps, speed limits, and the designated route. 
The planner is tasked with generating a trajectory for the subsequent 8 seconds. It is essential to note that, unless otherwise stated, we employ the unaltered trajectory output from the planner in this paper. We intentionally avoid incorporating performance-enhancing techniques, such as rule-based emergency stops or post-optimization, to assess the planner's inherent performance.

\paragraph{nuPlan}~\cite{caesar2021nuplan} is a large-scale closed-loop ML-based planning benchmark for autonomous vehicles. The dataset encompasses 1300 hours of recorded driving data collected from four urban centers, segmented into 75 scenario types using automated labeling tools. 

\paragraph{Simulation.} 
We use nuPlan's closed-loop simulator as our simulation environment. 
Each simulation entails a 15-second rollout at a rate of 10 Hz. 
It employs an LQR controller for trajectory tracking, while the control commands are utilized to update the state of the autonomous vehicle through an internal kinematic bicycle model. 
The behavior of background traffic varies based on the simulation mode, which can be non-reactive (log-replay) or reactive. 

\begin{table*}
\vspace{6pt}
\begin{center}
\setlength{\tabcolsep}{10pt}
\renewcommand{\arraystretch}{1.2}
\small
\begin{tabular}{y{50}y{85}|x{35}x{35}x{35}|x{35}x{35}x{35}}
\toprule
\multicolumn{2}{c}{Models} & \multicolumn{3}{c}{Test14-random} & \multicolumn{3}{c}{Test14-hard} \\ \midrule
Input feature & \multicolumn{1}{l|}{Variants} & OLS & NR-CLS & \multicolumn{1}{c|}{R-CLS} & OLS & NR-CLS & R-CLS \\ \midrule
\multirow{2}{*}{w/ history} & \multicolumn{1}{l|}{shared encoder} & 90.20 & 56.50 & \multicolumn{1}{c|}{56.28} & \textbf{88.25} & 48.60 & 51.32 \\
 & \multicolumn{1}{l|}{seperate encoder} & \textbf{90.28} & 61.02 & \multicolumn{1}{c|}{59.85} & 86.77 & 51.98 & 49.34 \\ \midrule
\multirow{5}{*}{w/o history} & \multicolumn{1}{l|}{state3 (x, y, yaw)} & 81.13 & \textbf{85.99} & \multicolumn{1}{c|}{\textbf{79.38}} & 71.43 & 68.44 & \textbf{63.14} \\
 & \multicolumn{1}{l|}{state4 (+vel.)} & 86.42 & 81.32 & \multicolumn{1}{c|}{75.75} & 82.30 & 68.15 & 62.51 \\
 & \multicolumn{1}{l|}{staet5 (+acc.)} & 87.71 & 81.76 & \multicolumn{1}{c|}{74.51} & 84.54 & \textbf{68.67} & 54.91 \\
 & \multicolumn{1}{l|}{state6 (+steer)} & 88.45 & 83.32 & \multicolumn{1}{c|}{77.52} & 85.93 & 65.15 & 55.99 \\ \bottomrule
\end{tabular}
\end{center}
\vspace{0.5em}
\caption{Results of different input features. For models with history input, ``shared" and ``separate" encoders indicate whether both the agent and ego vehicles utilize a shared history encoder or distinct ones. For models without history input, $+$ refers to the inclusion of an additional state variable compared to the preceding model in the table. \underline{Higher values} indicate better performance for all metrics, with the best metric highlighted in \textbf{bold}.}
\label{tab:ego_features}
\vspace{-0.5em}
\end{table*}

\paragraph{Metrics.} 
We employ the official evaluation metrics provided by nuPlan, which include the open-loop score (OLS), non-reactive closed-loop score (NR-CLS), and reactive closed-loop score (R-CLS). 
R-CLS and NR-CLS share identical calculation methodologies, differing only in that R-CLS incorporates background traffic control via an Intelligent Driver Model (IDM)~\cite{treiber2000idm} during simulations. 
The closed-loop score is a comprehensive composite score, achieved through a weighted combination of factors such as traffic rule adherence, human driving resemblance, vehicle dynamics, goal attainment, and other metrics specific to the scenario. The score scales from 0 to 100.  
For a detailed description and calculation of the metrics, please refer to \cite{nuplan2023metrics}.

\paragraph{Baseline.}
As a baseline, we have adapted the motion-forecasting backbone model from our prior work~\cite{cheng2023forecast} to address the planning task. Figure \ref{fig:baseline_model} provides a concise overview of the baseline model.
Despite its simplicity, the architecture primarily comprises multiple Transformer encoders~\cite{vaswani2017attention}, demonstrating significant modeling capacity. We direct interested readers to the code base for details.

\paragraph{Benchmark.} 
For all experiments, we standardize the data split for training and evaluation. For the training phase, we utilize all 75 scenario types in the nuPlan training set, limiting the total number of scenarios to 1M frames. For the evaluation phase, we employ 14 scenario types specified by the nuPlan Planning Challenge, each comprising 20 scenarios. We examine two scenario selection schemes: (1) \textbf{Test14-random}: scenarios are randomly sampled from each type and fixed after selection, and (2) \textbf{Test14-hard}: in order to investigate the planner's performance on long-tail scenarios, we execute 100 scenarios of each type using a state-of-the-art rule-based planner (PDM-Closed~\cite{Dauner2023CORL}), subsequently selecting the 20 least-performing scenarios of each type. Example scenarios can be found on the project page.  As the online leaderboard submission is closed, all evaluations are conducted on the nuPlan public test set.

\subsection{Input feature makes a difference}
This section aims to address the following questions: 
(1) \textit{Is historical motion data essential for planning?} 
(2) \textit{If not, do all current states of autonomous vehicles contribute to improving the planner's performance?} 
To address these inquiries, we conducted an investigation involving two sets of variants derived from our baseline model. 
The results on Test14-random and Test14-hard benchmarks are presented in Table \ref{tab:ego_features}.
Among the two historical variants, one shares its history encoder with other traffic agents, while the other employs a distinct history encoder for the ego vehicle's past motion. 
In the case of state-only models, we scrutinized several pivotal state variables essential for conventional planners, encompassing vehicle pose, velocity, acceleration, and steering angle. 
Based on the experimental results, we have the following findings:

\begin{figure}
    \centering
    \includegraphics[width=1.0\linewidth]{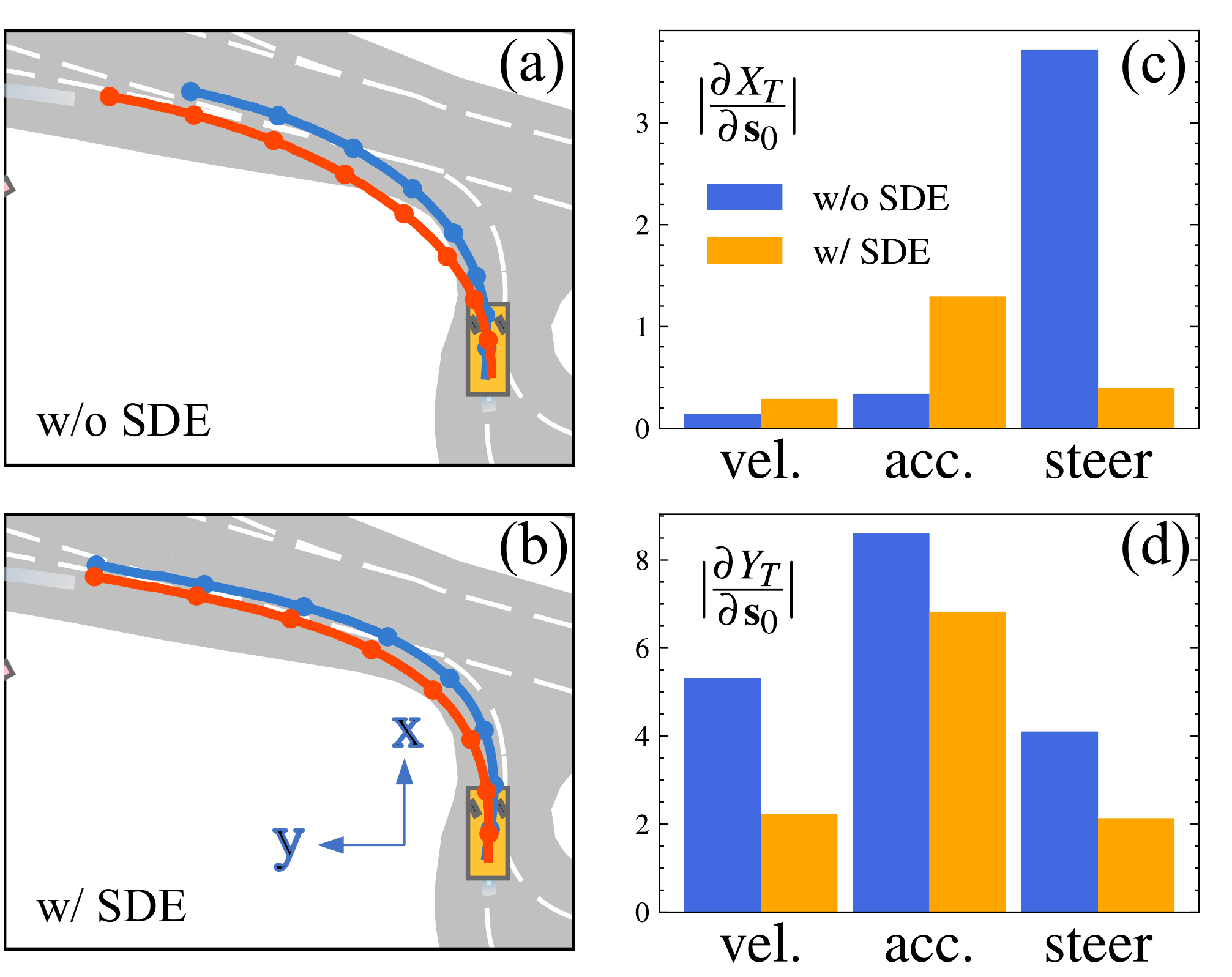}
    \vspace{-1em}
    \caption{The left side shows the planning trajectory of the \textit{state6} model by adjusting AV's steering angle from \textcolor{blue}{0.15} to \textcolor{red}{0.5} rad. The right side illustrates the magnitude of the gradient concerning the trajectory endpoint's position in relation to the AV's kinematic states.}
    \label{fig:left_turn_case}
\end{figure}

\begin{figure}
    \centering
    \includegraphics[width=0.75\linewidth]{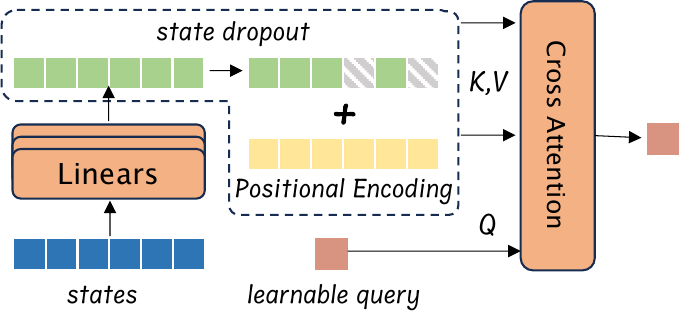}
    \vspace{-1em}
    \caption{Illustration of the attention-based state dropout encoder.}
    \label{fig:state_dropout}
\vspace{-1em}
\end{figure}

\paragraph{History is not necessary.} 
While models incorporating historical motion data exhibit superior off-policy evaluation performance (OLS), they manifest significantly poorer performance in closed-loop metrics compared to state-based models. 
This phenomenon may attributed to the well-established ``copycat'' problem~\cite{wen2020fighting} or learning shortcuts~\cite{geirhos2020shortcut}, wherein the planner relies on extrapolation from historical data without a comprehensive grasp of the underlying causal factors. 
Furthermore, the advantage in open-loop performance of historical models diminishes rapidly as the number of states increases in state-only models.  
Therefore, we conclude that history motions are not necessary for planning models. 

\begin{table*}[h!]
\begin{center}
\setlength{\tabcolsep}{5pt}
\renewcommand{\arraystretch}{1.2}
\small
\begin{tabular}{y{25}x{30}y{50}y{60}x{60}y{50}y{60}y{60}}
\toprule
 &  & \multicolumn{3}{c}{Test14-random} & \multicolumn{3}{c}{Test14-hard} \\ \midrule
 & \multicolumn{1}{c|}{SDE} & OLS & NR-CLS & \multicolumn{1}{l|}{R-CLS} & OLS & NR-CLS & R-CLS \\ \midrule
\fadedtext{staet3} & \multicolumn{1}{c|}{-} & \fadedtext{81.13} & \fadedtext{85.99} & \multicolumn{1}{l|}{\fadedtext{79.38}} & \fadedtext{71.43} & \fadedtext{68.44} & \fadedtext{63.14} \\ \midrule
\multirow{2}{*}{state5} & \multicolumn{1}{c|}{\xmark} & 87.71 & 81.76 & \multicolumn{1}{l|}{74.51} & 84.54 & 68.67 & 54.91 \\
 & \multicolumn{1}{c|}{\cmark} & 88.80 \textcolor{blue}{(+1.09)} & 86.73 \textcolor{blue}{(+4.97)} & \multicolumn{1}{l|}{75.75 \textcolor{blue}{(+1.24)}} & 84.29 \textcolor{red}{(-0.25)} & 71.28 \textcolor{blue}{(+2.61)} & 61.88 \textcolor{blue}{(+6.97)} \\ \midrule
\multirow{2}{*}{state6} & \multicolumn{1}{c|}{\xmark} & 88.55 & 83.19 & \multicolumn{1}{l|}{74.79} & 85.89 & 67.57 & 58.99 \\
 & \multicolumn{1}{c|}{\cmark} & 87.07 \textcolor{red}{(-1.48)} & 86.48 \textcolor{blue}{(+3.29)} & \multicolumn{1}{l|}{80.59 \textcolor{blue}{(+5.80)}} & 83.32 \textcolor{red}{(-2.57)} & 72.68 \textcolor{blue}{(+5.11)} & 61.70 \textcolor{blue}{(+2.71)} \\ \bottomrule
\end{tabular}
\end{center}
\vspace{0.5em}
\caption{Experimental results of the state dropout encoder (SDE) on Test14-random and Test14-hard benchmark. Models with SDE gain significant improvements on CLS while maintaining high performance on OLS.}
\label{tab:state_dropout}
\vspace{-1em}
\end{table*}

\paragraph{Shortcut learning in kinematic states.} 
Kinematic states, such as velocity and acceleration, serve as vital initial boundary conditions for ensuring safety and comfort in trajectory planning. 
Nevertheless, we are surprised to find that the \textit{state3} model, which exclusively relies on the autonomous vehicle's (AV) pose (comprising position and heading), significantly outperforms other models incorporating kinematic states in terms of CLS metrics. 
To gain deeper insights into this phenomenon, we study a left-turn case of the \textit{state6} model. 
As depicted in Figure \ref{fig:left_turn_case}\textcolor{red}{a}, the model generates an undesired off-road trajectory when changing the steering angle from 0.15 (blue) to 0.5 (red) rad. 
We hypothesize the model still learns false correlation from the kinematics even without the present of the past observation. 

\paragraph{State dropout encoder.}
To confront our assumption, we propose an attention-based state dropout encoder (SDE), as shown in Figure \ref{fig:state_dropout}. Each state variable undergoes individual embedding through a linear layer before being combined with positional encoding. A learnable query aggregates state embeddings through a cross-attention module. During training, each embedded state (except position and heading) token will be dropped with a certain probability. The encoder compels the model to unveil the root causes of behaviors by imposing partial constraints on its access to auxiliary information. Meanwhile, the model can enhance its planning capabilities when kinematic attributes are accessible. 
We implement the state dropout encoder in the \textit{state5} and \textit{state6} models, and the results are depicted in Table \ref{tab:state_dropout}. The results indicate that the utilization of SDE significantly enhances the closed-loop performance of the models. Importantly, when compared to \textit{state3}, \textit{state5} and \textit{state6} models augmented with SDE exhibit not only improved closed-loop score but also substantially higher open-loop score, providing compelling evidence for the efficacy of SDE. We point out that \textit{state3} model is fundamentally ambiguous as it loses all kinematic information (supported by its poor OLS performance). Figure 
\ref{fig:left_turn_case}(a)(b) displays the comparative planning results of \textit{state6} model, while Figure \ref{fig:left_turn_case}(c)(d) presents comparative results for the magnitude of the gradient of the endpoint's position $(X_T, Y_T)$ w.r.t. the initial kinematic states $s_0$. The results demonstrate that the model employing SDE is less sensitive to variations in kinematic states, resulting in more resilient planning outcomes. 

\begin{figure}
    \centering
    \includegraphics[width=0.75\linewidth]{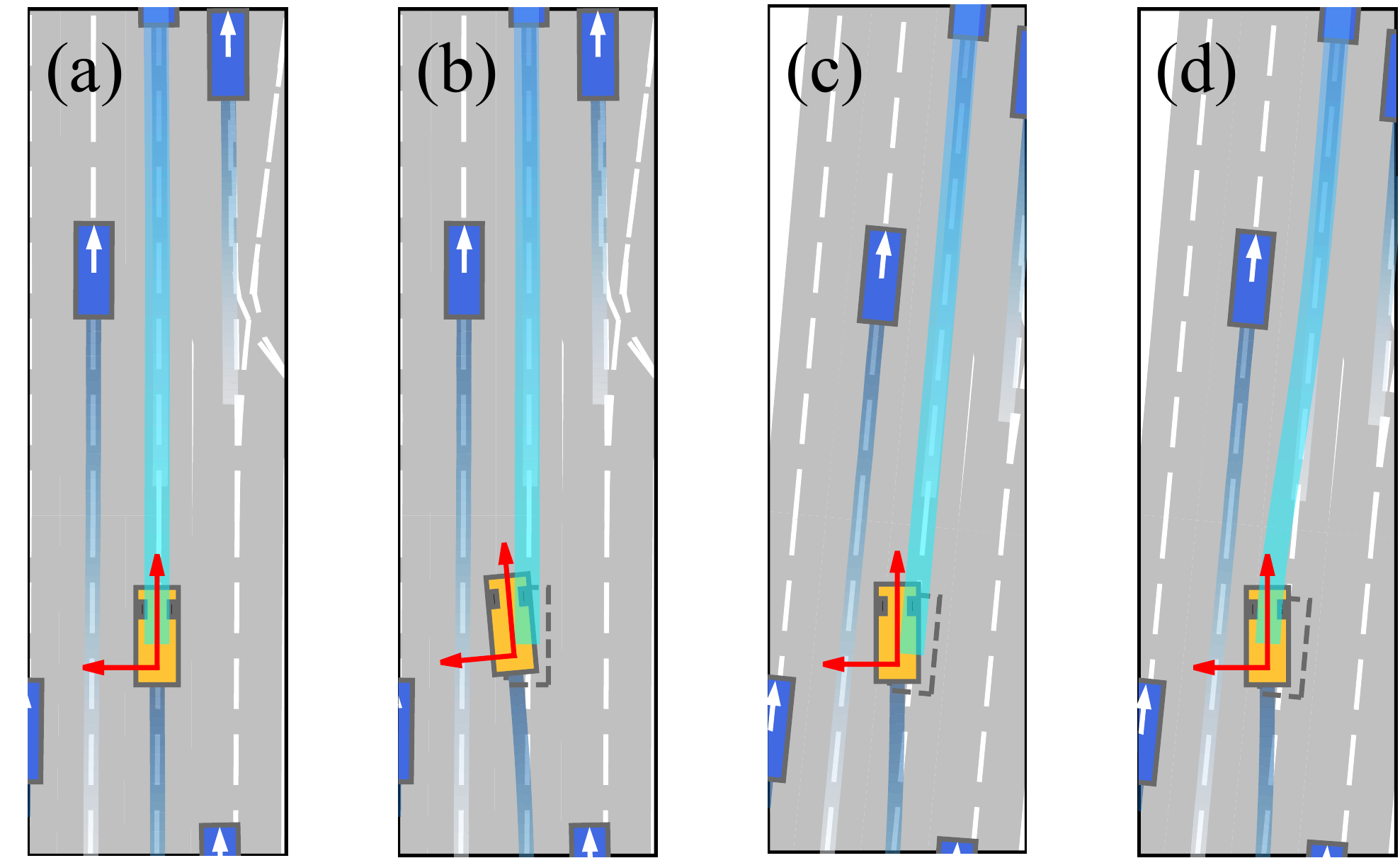}
    \vspace{-0.5em}
    \caption{(a) The original scenario. (b) Random noise is added to the AV's current state and history motion is smoothed. (c) The coordinates of the scenario are re-normalized based on the perturbed position of the AV. (d) A corrected future trajectory is generated using constrained nonlinear optimization. }
    \label{fig:aug_vis}
\end{figure}

\begin{table}[!h]
\begin{center}
\setlength{\tabcolsep}{5pt}
\renewcommand{\arraystretch}{1.2}
\small
\begin{tabular}{y{25}x{10}x{15}x{15}|x{30}x{35}x{30}}
\toprule
 & P & RN & FC & OLS & NR-CLS & R-CLS \\ \midrule
\multirow{2}{*}{history} & \xmark & \xmark & \xmark & 88.99 & 65.84 & 65.58 \\
 & \cmark & \cmark & \cmark & 89.94 & 65.14 & 66.03 \\ \midrule
\multirow{4}{*}{state3} & \xmark & \xmark & \xmark & 78.92 & 71.86 & 70.87 \\
 & \cmark & \xmark & \xmark & 80.85 & 74.28 & 71.69 \\
 & \cmark & \cmark & \xmark & 81.13 & 85.99 & 79.38 \\
 & \cmark & \cmark & \cmark & 79.28 & 81.35 & 76.60 \\ \midrule
\multirow{4}{*}{state5} & \xmark & \xmark & \xmark & 88.44 & 80.67 & 74.50 \\
 & \cmark & \xmark & \xmark & 89.20 & 79.85 & 72.43 \\
 & \cmark & \cmark & \xmark & 87.71 & 81.76 & 74.51 \\
 & \cmark & \cmark & \cmark & 86.43 & 82.10 & 74.71 \\ \midrule
\multicolumn{1}{c}{\multirow{4}{*}{\begin{tabular}[c]{@{}c@{}}state6\\ +SDE\end{tabular}}} & \xmark & \xmark & \xmark & 88.33 & 77.28 & 74.10 \\
\multicolumn{1}{c}{} & \cmark & \xmark & \xmark & 87.71 & 77.70 & 75.18 \\
\multicolumn{1}{c}{} & \cmark & \cmark & \xmark & 87.07 & 86.48 & 80.59 \\
\multicolumn{1}{c}{} & \cmark & \cmark & \cmark & 85.50 & 82.95 & 76.09 \\ \bottomrule
\end{tabular}
\end{center}
\caption{Results of different augmentation and normalization combinations on \textbf{Test14-random} benchmark. P: \textbf{P}erturbation; RN: \textbf{R}e-\textbf{N}ormalization; FC: \textbf{F}uture \textbf{C}orrection.}
\label{tab:aug}
\vspace{-2em}
\end{table}

\subsection{Data augmentation and normalization}

Data augmentation is a common practice for IL-based models to learn how to recover from deviations. 
In this section, we conduct comprehensive experiments on different data augmentation techniques, aiming to explore effective strategies to mitigate compounding errors. Different augmentation strategies are displayed in Figure \ref{fig:aug_vis}. In Figure \ref{fig:aug_vis}(a), an example driving scenario is depicted, with all coordinates normalized relative to the autonomous vehicle's center. In Figure \ref{fig:aug_vis}(b), randomly sampled noise (perturbation) is added to the AV's current state, and its history states are smoothed accordingly. In Figure \ref{fig:aug_vis}(c), it is demonstrated that the scenario's coordinates are re-normalized with respect to the autonomous vehicle's center after perturbation. Figure \ref{fig:aug_vis}(d) showcases the generation of a rectified future trajectory through nonlinear optimization. Essentially, both strategies depicted in Figure \ref{fig:aug_vis}(b) and \ref{fig:aug_vis}(d) serve the common objective of guiding the vehicle back to the expert trajectory. 

Table \ref{tab:aug} displays the outcomes of experiments conducted with various augmentation strategies on four model variants. Based on these results, the following findings emerge: (1) In the case of the \textit{history} and \textit{state5} models, none of the data augmentations exhibit substantial enhancements. We postulate that the primary challenge faced by these two models is the issue of causal confusion, \ie extrapolation from either historical or kinematic states. (2) For \textit{state3} and \textit{state6+SDE} models, perturbation is of great importance, but only works with proper normalization. For example, \textit{state3} model's NR-CLS score boosts from 71.86 to 85.99 with perturbation and re-normalization, which is much higher than solely using perturbation (74.28). This implies it is important to keep the data distribution close between training and testing. (3) Providing a corrected guiding future trajectory does not serve a positive effect. One possible reason is that the manually generated trajectory does not align with the expert's trajectory distribution. Directly using the expert trajectory as supervision is a more effective choice as it keeps the original distribution, and small deviations can be easily fixed by the tracker. 

\begin{figure}
    \centering
    \includegraphics[width=0.9\linewidth]{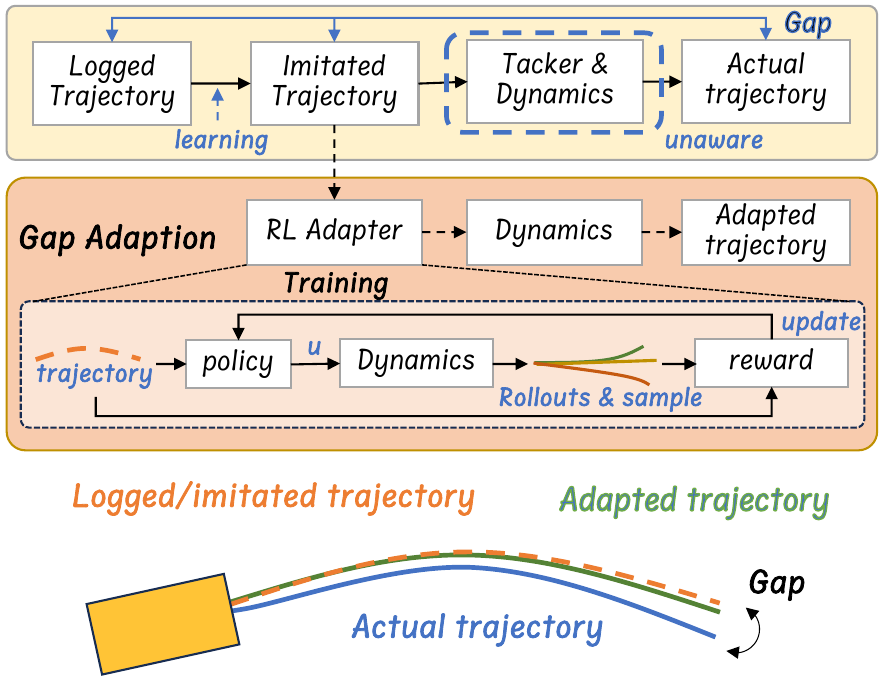}
    \caption{Illustration of the imitation gap and the proposed RL adapter. }
    \label{fig:imitatoin_gap}
\end{figure}

\begin{table}[]
\begin{center}
\setlength{\tabcolsep}{5pt}
\renewcommand{\arraystretch}{1.2}
\small
\begin{tabular}{y{60}x{35}x{30}|x{35}x{30}}
\toprule
                                      & \multicolumn{2}{c}{Test14-random}   & \multicolumn{2}{c}{Test14-hard} \\ \midrule
\multicolumn{1}{l|}{Log-replay +}          & NR-CLS & \multicolumn{1}{c|}{R-CLS} & NR-CLS          & R-CLS         \\ \midrule
\multicolumn{1}{l|}{\fadedtext{Perfect tracking}} & \fadedtext{96.63}  & \multicolumn{1}{c|}{\fadedtext{77.38}} & \fadedtext{91.61}           & \fadedtext{71.34}         \\ \midrule
\multicolumn{1}{l|}{LQR}              & 94.03  & \multicolumn{1}{c|}{75.86} & 85.96           & 68.80          \\
\multicolumn{1}{l|}{RL Adapter}       & 96.3   & \multicolumn{1}{c|}{77.13} & 91.65           & 71.62         \\ \bottomrule
\end{tabular}
\end{center}
\vspace{0.5em}
\caption{Exprimental results of the log-replay planner (perfect imitation) with different trackers on Test14-random and Test14-hard benchmarks. LQR is the default tracker used by the nuPlan benchmark and the RL adapter is our proposed method to address the imitation gap.}
\label{tab:il_gap}
\vspace{-1em}
\end{table}

\subsection{The hidden imitation gap}

\paragraph{The imitation gap.} Within the most popular imitation learning frameworks, models imitate the logged expert's footprints from the dataset. We argue that this learning framework gives rise to a concealed gap in imitation, potentially leading to notable performance degradation. As illustrated in Figure \ref{fig:imitatoin_gap}, the recorded trajectory, commonly known as the expert trajectory, serves as the ground truth during the training of the imitation-based planner. The generated imitated trajectory is subsequently processed by the downstream tracker and the underlying system dynamics, yielding the final trajectory of the AV. Nevertheless, owing to the lack of knowledge about the tracker and dynamics during training, the actual trajectory may substantially deviate from the recorded trajectory, even when imitation is flawless. This assertion finds support in the experimental findings presented in Table \ref{tab:il_gap}. Notably, the NR-CLS of the \textit{Log-replay + LQR} method on Test14-hard exhibits a significant decrease of $5.65$ in comparison to perfect tracking.

\begin{table}
\begin{center}
\setlength{\tabcolsep}{5pt}
\renewcommand{\arraystretch}{1.1}
\small
\begin{tabular}{y{70}x{70}x{60}}
\toprule
Reward Term       & Expression & Weight \\ \midrule
Position Tracking & $e^{-15||\mathbf{p}_{xy}-\mathbf{p}^*_{xy}||}$ &  1.0     \\
Action            & $||\mathbf{u}||^2$ & -0.01    \\
Action Rate       & $||\dot{\mathbf{u}}||^2$ & -0.1     \\
Lon. Acc. limit   & $\mathbb{1}({\dot{v}}>2.4)$ & -1    \\
Jerk limit        & $\mathbb{1}(||{\ddot{v}}||>4.0)$  & -1        \\
Yaw rate          & $\mathbb{1}(||{\dot{\theta}}||>0.95)$ & -0.5        \\ \bottomrule
\end{tabular}
\end{center}
\vspace{0.5em}
\caption{The reward terms and expression of the RL adapter. Action $\mathbf{u}$ contains acceleration and steering rate. $v$ and $\theta$ refers to the longitudinal and heading angle of the AV.}
\label{tab:rewards}
\vspace{-1em}
\end{table}

\begin{table*}[]
\vspace{6pt}
\begin{center}
\setlength{\tabcolsep}{10pt}
\renewcommand{\arraystretch}{1.2}
\small
\begin{tabular}{y{58}y{70}|x{25}x{33}x{30}|x{25}x{33}x{30}x{30}}
\toprule
\multicolumn{2}{c}{Planners} & \multicolumn{3}{c}{Test14-random} & \multicolumn{3}{c}{Test14-hard} &  \\ \midrule
Type & \multicolumn{1}{l|}{Method} & OLS & NR-CLS & \multicolumn{1}{c|}{R-CLS} & OLS & NR-CLS & \multicolumn{1}{c|}{R-CLS} & Time(ms) \\ \midrule
\fadedtext{Expert} & \multicolumn{1}{l|}{\fadedtext{Log-replay}} & \fadedtext{100.0} & \fadedtext{94.03} & \multicolumn{1}{c|}{\fadedtext{75.86}} & \fadedtext{100.0} & \fadedtext{85.96} & \multicolumn{1}{c|}{\fadedtext{68.80}} & \fadedtext{-} \\ \midrule
\multirow{2}{*}{Rule-based} & \multicolumn{1}{l|}{IDM~\cite{treiber2000idm}} & 34.15 & 70.39 & \multicolumn{1}{c|}{72.42} & 20.07 & 56.16 & \multicolumn{1}{c|}{62.26} & 32 \\
 & \multicolumn{1}{l|}{PDM-Closed~\cite{Dauner2023CORL}} & 46.32 & 90.05 & \multicolumn{1}{c|}{\textbf{91.64}} & 26.43 & 65.07 & \multicolumn{1}{c|}{75.18} & 140 \\ \midrule
\multirow{2}{*}{Hybrid$^\dagger$} & \multicolumn{1}{l|}{GameFormer~\cite{huang2023gameformer}} & 79.35 & 80.80 & \multicolumn{1}{c|}{79.31} & 75.27 & 66.59 & \multicolumn{1}{c|}{68.83} & 443 \\
 & \multicolumn{1}{l|}{PDM-Hybrid~\cite{Dauner2023CORL}} & 82.21 & \textbf{90.20} & \multicolumn{1}{c|}{91.56} & 73.81 & 65.95 & \multicolumn{1}{c|}{\textbf{75.79}} & 152 \\ \midrule
\multirow{5}{*}{Learning-based} & \multicolumn{1}{l|}{RasterModel~\cite{caesar2021nuplan}} & 62.93 & 69.66 & \multicolumn{1}{c|}{67.54} & 52.4 & 49.47 & \multicolumn{1}{c|}{52.16} & 82 \\
 & \multicolumn{1}{l|}{UrbanDriver~\cite{scheel2022urban}} & 82.44 & 63.27 & \multicolumn{1}{c|}{61.02} & 76.9 & 51.54 & \multicolumn{1}{c|}{49.07} & 124 \\
 & \multicolumn{1}{l|}{GC-PGP~\cite{hallgarten2023gc-pgp}} & 77.33 & 55.99 & \multicolumn{1}{c|}{51.39} & 73.78 & 43.22 & \multicolumn{1}{c|}{39.63} & 160 \\
 & \multicolumn{1}{l|}{PDM-Open~\cite{Dauner2023CORL}} & 84.14 & 52.80 & \multicolumn{1}{c|}{57.23} & 79.06 & 33.51 & \multicolumn{1}{c|}{35.83} & 101 \\
 & \multicolumn{1}{l|}{PlanTF (Ours)} & \textbf{87.07} & 86.48 & \multicolumn{1}{c|}{80.59} & \textbf{83.32} & \textbf{72.68} & \multicolumn{1}{c|}{61.7} & 155 \\ \bottomrule
\end{tabular}
\end{center}
\caption{Comparison with state-of-the-arts. The runtime includes feature extraction and model inference based on Python code.\\\textit{\footnotesize $^\dagger$ indicates these methods' final output trajectory relies on rule-based strategies or post-optimization.}}
\label{tab:sota}
\vspace{-1em}
\end{table*}

\paragraph{RL Adapter.} One possible solution is to directly imitate the control command rather than the trajectory points. Nevertheless, this approach is heavily reliant on the specific vehicle model, making it less generalizable and interpretable than the trajectory-based method. An alternative approach involves incorporating a differentiable kinematic model into the trajectory decoder~\cite{cui2020deep,zhou2021exploring}. However, the kinematic model is often oversimplified to ensure differentiability. To tackle this challenge, we introduce a reinforcement learning-based trajectory adapter (RL Adapter) designed to bridge this gap. The RL Adapter transforms the imitated trajectory into the relevant control commands while accounting for the underlying dynamics. The benefits are two-folds. First, it can adapt to various vehicle models without retraining the planner. Second, it imposes no constraints on the vehicle model and remains compatible with non-differentiable vehicle models (\eg high-fidelity real vehicle dynamics model). The training process of the adapter is displayed in Figure \ref{fig:imitatoin_gap} and the rewards are shown in Table \ref{tab:rewards}. We use PPO~\cite{schulman2017proximal} for policy optimization and the training finishes in 80K steps with a learning rate of 1e-3. As depicted in Table \ref{tab:il_gap}, the RL Adapter performs similarly to perfect tracking, highlighting its capacity to bridge the imitation gap. We notice that it can be integrated into the training process of the planner and leave this as future work.

\section{Comparison to State of the Art}

\paragraph{Implementation details.} 
Integrating our findings, we propose a fully learning-based baseline planning model called Planning Transformer (\textbf{PlanTF}). Specifically, we employ the \textit{state6} model, incorporating a state attention dropout encoder with a dropout rate of 0.75. During training, we apply state perturbation with a probability of 0.5. The model is trained using a batch size of 128 and a weight decay of 1e-4 for 25 epochs. The initial learning rate is set to 1e-3, decaying to zero in a cosine manner.

\paragraph{Methods.}
We compare PlanTF's performance with several state-of-the-art planners. 
(1) \textbf{RasterModel} is a CNN-based planner provided in~\cite{caesar2021nuplan}. 
(2) \textbf{UrbanDriver}~\cite{scheel2022urban} is a vectorized planner based on PointNet-based polyline encoders and Transformer. Here we use its open-loop re-implementation and history perturbation is employed during training. 
(3) \textbf{GameFormer}~\cite{huang2023gameformer} is a DETR-like interactive prediction and planning framework based on the level-k game, which incorporates a post-optimizer to generate the final trajectory.
(4) \textbf{PDM*}~\cite{Dauner2023CORL} is the winning solution of the 2023 nuPlan Planning Challenge. \textbf{PDM-Closed} is a purely rule-based approach that ensembles the IDM~\cite{treiber2000idm} with different hyperparameters. \textbf{PDM-Hybrid} is a variant of PDM-closed that adds an offset predictor to improve its open-loop prediction performance. \textbf{PDM-Open} is the pure learning component without the IDM-based planner. Results are reproduced using their publicly available code and trained on our standard 1M data split.

\paragraph{Results.}
Table \ref{tab:sota} presents comparative results for the Test14-random and Test14-hard benchmarks. First, the proposed PlanTF significantly outperforms all other pure imitation-based methods across all metrics, particularly in terms of closed-loop performance. It is also the only learning-based method that surpasses the widely recognized IDM, highlighting the importance of proper design in IL. Second, when compared to rule-based and hybrid methods, PlanTF delivers outstanding OLS while maintaining highly competitive CLS, without the need for any tricky hand-crafted rules or strategies. Notably, our approach achieves the highest NR-CLS on the Test14-hard benchmark, indicating that although rule-based methods perform well in ordinary scenarios (Test14-random), they struggle to generalize in long-tail situations (Test14-hard). In contrast, PlanTF demonstrates stronger generalization capabilities. 

\section{Conclusion}

In this study, we systematically examine several crucial design aspects of imitation-based planners by utilizing the standardized nuPlan benchmark. Our findings reveal that catastrophic shortcut learning generally occurs for input features, such as historical motions and single-frame kinematic states. This leads to the unexpected outcome that planning solely based on the AV's current position results in superior closed-loop performance. To mitigate this issue, we introduce a straightforward attention-based state dropout encoder (SDE) that effectively addresses the shortcut learning problem. With the implementation of SDE, the \textit{state6} model achieves the best overall performance. Data augmentation is another significant factor in imitation-based planners. Our results demonstrate that perturbation is vital for reducing compounding errors, but only effective with appropriate feature normalization. Furthermore, we observe that the original expert trajectory remains a reliable training ground truth, even when subjected to perturbation. In addition to these findings, we identify a neglected imitation gap caused by the model's lack of awareness of the underlying system dynamics, which considerably impacts the planner's performance. To rectify this issue, we propose a reinforcement learning-based adapter. By incorporating our findings, the proposed purely learning-based baseline model, PlanTF, demonstrates impressive performance compared to state-of-the-art approaches and is on par with methods that employ intricate rule-based strategies or post-optimization. This highlights the importance of proper design choices for imitation learning-based planners.

\paragraph{Limitation and future work. } 
Despite pushing the boundaries of pure imitation-based planners, our method is constrained by the fundamental mismatch between \textit{open-loop training} and \textit{closed-loop testing}. Incorporating closed-loop information and system dynamics into the training process constitutes our future research direction.

\appendix

\paragraph{Additional results on Val14 benchmark.} We present the comparative results (Table. \ref{tab:val14}) on the \textbf{Val14}~\cite{Dauner2023CORL} benchmark. \textbf{Val14} contains 1180 scenarios from 14 scenario types.

\begin{table}[h]
\vspace{6pt}
\begin{center}
\setlength{\tabcolsep}{7pt}
\renewcommand{\arraystretch}{1.2}
\small
\begin{tabular}{y{70}|x{30}x{35}x{30}}
\toprule
Method       & OLS & NR-CLS & R-CLS \\ \midrule
\fadedtext{Log-replay}   & \fadedtext{100} & \fadedtext{94}     & \fadedtext{80}    \\
IDM~\cite{treiber2000idm}          & 38  & 77     & 76    \\
GC-PGP~\cite{hallgarten2023gc-pgp}       & 82  & 57     & 54    \\
PlanCNN~\cite{caesar2021nuplan}      & 64  & 73     & 72    \\
PDM-Hybrid~\cite{Dauner2023CORL}   & 84  & 93     & 92    \\ \midrule
PlanTF (Ours) & 89.18    & 84.83  & 76.78 \\ \bottomrule
\end{tabular}
\end{center}
\vspace{0.5em}
\caption{Comparision to SOTAs on the Val14 benchmark. The results of other methods are taken from~\cite{Dauner2023CORL}.}
\label{tab:val14}
\end{table}

\paragraph{Ablation on the state dropout rate.} Table \ref{tab:abl_dropout} shows the ablation study on different dropout rate the of \textit{state6+SDE} model.

\begin{table}[h]
\vspace{6pt}
\begin{center}
\setlength{\tabcolsep}{6pt}
\renewcommand{\arraystretch}{1.2}
\small
\begin{tabular}{y{30}|x{30}|x{35}x{35}x{30}}
\toprule
Model                   & \multicolumn{1}{l|}{dropout} & OLS   & NR-CLS & R-CLS \\ \midrule
\multirow{4}{*}{state6} & -                          & 88.33 & 77.28  & 74.10 \\
                        & 0.25                         & 89.11 & 81.70  & 78.44 \\
                        & 0.50                         & 89.12 & 83.71  & 77.52 \\
                        & 0.75                         & 87.07 & 86.48  & 80.59 \\ \bottomrule
\end{tabular}
\end{center}
\vspace{0.5em}
\caption{Ablation study on the state dropout rate of the SDE.}
\label{tab:abl_dropout}
\end{table}


\bibliographystyle{IEEEtran}
\bibliography{IEEEabrv,ref}

\end{document}